\documentclass[conference]{IEEEtran}
\usepackage{caption}
\usepackage{algorithm}
\usepackage{algpseudocode}
\usepackage{float}
\usepackage{cite}
\usepackage{graphicx}
\usepackage{amsmath}
\usepackage{booktabs}
\usepackage[hidelinks]{hyperref}

\begin{document}

\title{Hybrid JIT--CUDA Graph Optimization for Low-Latency Large Language Model Inference}

\author{
\IEEEauthorblockN{Divakar Kumar Yadav}
\IEEEauthorblockA{Department of Computer Science\\
University of Wisconsin-Milwaukee\\
Milwaukee, WI, USA\\
dyadav@uwm.edu}
\and
\IEEEauthorblockN{Tian Zhao}
\IEEEauthorblockA{Department of Computer Science\\
University of Wisconsin-Milwaukee\\
Milwaukee, WI, USA\\
tzhao@uwm.edu}
}


\maketitle

\begin{abstract}
Large Language Models (LLMs) have achieved strong performance across natural language and multimodal tasks, yet their practical deployment remains constrained by inference latency and kernel launch overhead, particularly in interactive, short-sequence settings. This paper presents a hybrid runtime framework that combines Just-In-Time (JIT) compilation with CUDA Graph execution to reduce launch overhead while preserving runtime flexibility during autoregressive decoding. The framework partitions transformer inference into static components executed via CUDA Graph replay and dynamic components handled through JIT-compiled kernels, enabling asynchronous graph capture and reuse across decoding steps.

We evaluate the proposed approach on LLaMA-2 7B using single-GPU, batch-size-one inference across prompt lengths from 10 to 500 tokens. Experimental results show that the hybrid runtime reduces Time-to-First-Token (TTFT) by up to 66.0\% and achieves lower P99 latency compared with TensorRT–LLM in this regime. These results indicate that hybrid JIT–CUDA Graph execution can effectively reduce inference latency and variance for short-sequence LLM workloads, making it a practical optimization strategy for latency-sensitive AI applications.
\end{abstract}


\section{Introduction}

Large Language Models (LLMs) such as GPT-4~\cite{brown2020language}, Claude~\cite{anthropic2024claude3},
Gemini~\cite{google2024gemini}, and Grok~\cite{xai2024grok}
have rapidly evolved into core inference engines for conversational, coding, and multimodal AI systems.
Despite their strong model capabilities, the real-time deployment of LLMs remains constrained by inference
latency, particularly in interactive settings where users generate and consume responses incrementally.
Prior work has shown that autoregressive decoding incurs substantial overhead from repeated GPU kernel
launches, synchronization events, and host--device coordination~\cite{narayanan2021efficient,rasley2020deepspeed,aminabadi2022deepspeed,kwon2023pagedattention}.
These costs accumulate across decoding steps and become especially visible in latency-sensitive inference scenarios.

To mitigate inference overhead, modern LLM runtimes employ a combination of compiler- and kernel-level
optimizations, including operator fusion~\cite{sarofeen2022nvfuser,tillet2019triton,li2023torchcompileinternals},
quantization~\cite{dettmers20228bit,frantar2023gptq}, and specialized attention kernels such as
FlashAttention~\cite{dao2022flashattention,dao2023flashattention2}. Production-oriented frameworks
such as TensorRT--LLM~\cite{nvidia2023tensorrtllm} and FasterTransformer~\cite{nvidia2023fastertransformer}
combine these techniques to improve throughput and reduce memory overhead. Nevertheless, even highly
optimized pipelines continue to incur non-trivial latency from repeated kernel dispatch and dynamic
tensor management during autoregressive decoding, particularly for small batch sizes and interactive use
cases.

\begin{figure*}[ht!]
\centering
\includegraphics[width=0.6\linewidth]{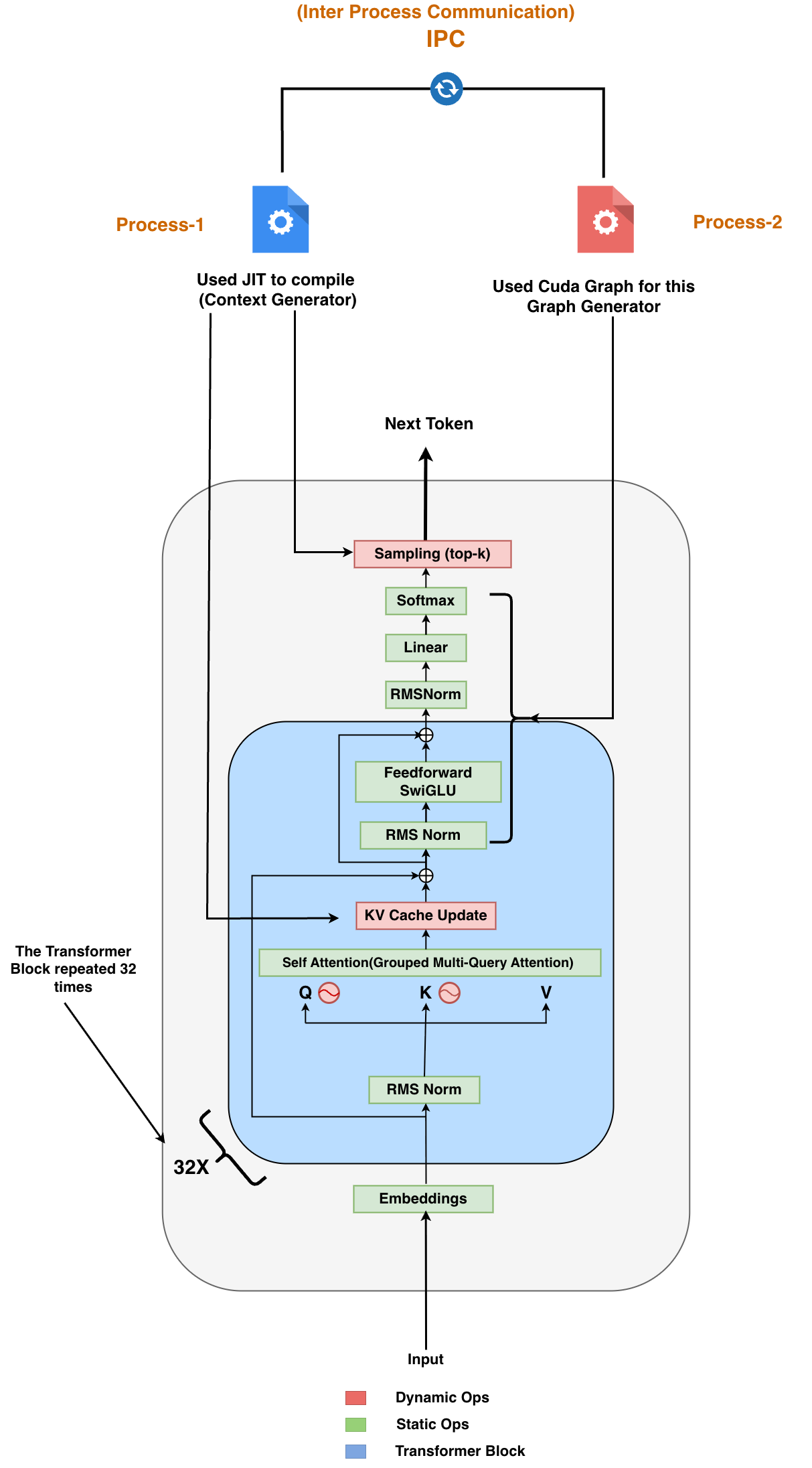}
\caption{High-level architecture of the Hybrid JIT--CUDA Graph Runtime.
Process~1 (JIT Context Generator) executes dynamic operations such as preprocessing and sampling, while
Process~2 (CUDA Graph Generator) executes static compute kernels through captured CUDA Graphs, coordinated
via inter-process communication (IPC).}
\label{fig:overview}
\end{figure*}

\subsection{Inference Length}
Interactive LLM applications, including conversational agents, code assistants, and decision-support
systems, are typically dominated by short-to-medium generation workloads in which responses are produced
incrementally and latency sensitivity is high. Public model documentation and large-scale observational
studies of deployed systems~\cite{anthropic2024claude3,chatterji2025chatgpt} consistently indicate that
many real-world interactions favor relatively limited response lengths rather than long-form generation.

From a systems perspective, this regime is particularly important because inference latency and variance
are most perceptible to end users during early decoding steps and moderate-length generations. Optimizing
execution for response lengths on the order of a few hundred tokens therefore addresses a common and
operationally relevant class of workloads, even when longer-context capabilities are available.

In this work, we focus on autoregressive inference up to 500 generated tokens as a representative window
for latency-sensitive evaluation. This choice reflects practical deployment considerations rather than an
assumption of universal workload distributions, and allows us to study kernel launch overhead, replay
behavior, and tail latency under realistic interactive conditions.

\subsection{Problem Context}
A primary bottleneck in LLM inference arises from the CPU-bound dispatch of numerous fine-grained GPU
kernels executed sequentially at each decoding step. Even frameworks that employ dynamic graph tracing
and compilation (e.g. \texttt{torch.compile}) cannot fully eliminate host-side coordination and Python
dispatch overhead~\cite{ramasamy2022torchdynamo}. CUDA Graphs~\cite{nvidia2024graphs,pytorchcudagraph}
provide a mechanism to pre-capture and replay GPU workloads with minimal host intervention, substantially
reducing launch overhead. However, CUDA Graphs require static tensor shapes and deterministic control
flow, making them incompatible with dynamic operations such as variable-length attention, cache updates,
and stochastic sampling commonly found in LLM inference.

Conversely, Just-In-Time (JIT) compilation~\cite{pytorch_jit,ansel2024pytorch} can accommodate dynamic
control flow and runtime-dependent tensor shapes, but JIT-executed kernels still incur per-launch overhead
and variability. As a result, existing approaches face a fundamental trade-off between static execution
efficiency and dynamic flexibility during autoregressive decoding.

\subsection{Contributions}
This work proposes a hybrid \textbf{JIT--CUDA Graph} runtime, illustrated in Fig.~\ref{fig:overview},
that bridges this trade-off by combining static CUDA Graph replay with dynamic JIT execution. Specifically,
we make the following contributions:
\begin{itemize}
  \item We partition the transformer inference pipeline into static components that are safe for CUDA
  Graph capture and dynamic components that require JIT execution.
  \item We introduce an asynchronous CUDA Graph generation mechanism that captures static subgraphs at
  multiple sequence lengths without blocking inference execution.
  \item We overlap JIT-executed dynamic operations with deterministic CUDA Graph replay to reduce kernel
  launch overhead and latency variance during autoregressive decoding.
  \item We demonstrate measurable reductions in Time-to-First-Token (TTFT) and tail latency for
  short-to-medium length LLM inference workloads on a single GPU.
\end{itemize}

The remainder of the paper is structured as follows: Section~\ref{sec:related-work} surveys related work; Section~\ref{sec:system} introduces
the hybrid framework and architecture; Section~\ref{sec:implementation} describes implementation details and fairness methodology;
Section~\ref{sec:results} presents empirical evaluation; Section~\ref{sec:discussion} discusses limitations; and Section~\ref{sec:conclusion} concludes with
future directions.

\section{Related Work}
\label{sec:related-work}

\subsection{Transformer Optimization}

Transformer-based architectures~\cite{vaswani2017attention,brown2020language,touvron2023llama,jiang2023mistral}
form the foundation of modern generative AI systems.
Scaling these models for efficient inference has motivated a rich ecosystem of GPU-optimized runtimes,
including Megatron-LM~\cite{shoeybi2019megatron},
DeepSpeed-Inference~\cite{aminabadi2022deepspeed,rasley2020deepspeed},
FasterTransformer~\cite{nvidia2023fastertransformer},
and TensorRT--LLM~\cite{nvidia2023tensorrtllm}.
These systems employ techniques such as mixed-precision execution, fused attention kernels,
and pipeline or tensor parallelism to improve throughput and reduce memory footprint.

While highly effective for batch-oriented and throughput-driven workloads, these runtimes are typically
designed around static or semi-static execution graphs. As a result, accommodating fine-grained dynamic
behavior -- such as variable sequence lengths, token-by-token autoregressive decoding, and stochastic
sampling -- often requires falling back to host-side coordination, which can introduce additional latency
and variance in interactive inference settings.

\subsection{Kernel Fusion}

Compiler-based approaches aim to reduce kernel launch overhead by fusing operator boundaries at compile
time or runtime. For example, nvFuser~\cite{sarofeen2022nvfuser}, TorchDynamo~\cite{ramasamy2022torchdynamo}, and the
\texttt{torch.compile} stack~\cite{ansel2024pytorch,li2023torchcompileinternals} transform Python-level
operator graphs into optimized CUDA kernels through a combination of ahead-of-time and just-in-time
compilation. Domain-specific systems such as Triton~\cite{tillet2019triton} and TVM~\cite{chen2018tvm}
further enable operator specialization and aggressive fusion for deep learning workloads.

Despite these advances, fused kernels are typically invoked through host-side scheduling and remain
sensitive to dynamic shapes and control flow. In autoregressive decoding, where kernel invocation patterns
and tensor shapes evolve at each step, residual launch overhead and compilation boundaries can still
dominate end-to-end latency.

\subsection{CUDA Graph Replay}

CUDA Graphs~\cite{nvidia2024graphs,pytorchcudagraph,gray2024constant} provide a mechanism to capture and
replay GPU workloads with minimal host involvement, substantially reducing kernel launch overhead.
Recent work on composable and reusable graph execution~\cite{ghosh2025} demonstrates that CUDA Graph replay
can achieve near-zero CPU overhead and highly stable execution for static tensor workloads.

However, CUDA Graph capture requires fixed tensor shapes, deterministic memory allocation, and static
control flow. These constraints make it difficult to directly capture graphs that include dynamic
operations such as variable-length attention, KV-cache updates, or stochastic sampling. Consequently,
existing CUDA Graph-based pipelines~\cite{lee2022cudaopt} are typically restricted to fixed-shape batches
or pre-tokenized inputs, limiting their applicability to fully dynamic autoregressive LLM inference.

\subsection{Hybrid Compilation}

Hybrid execution strategies combine static compilation with dynamic execution to balance performance and
flexibility. Systems such as XLA~\cite{xla2017tensorflow} and TensorFlow’s runtime fusion infrastructure
demonstrate that selectively compiling stable subgraphs while interpreting dynamic regions can improve
overall utilization. More recent studies~\cite{ghosh2025} explore integrating JIT-compiled kernels with
CUDA Graph capture in transformer workloads, reporting improved latency scaling under controlled settings.

Nevertheless, many existing hybrid approaches either require re-implementing model operators in C++,
assume globally static graph shapes, or target limited portions of the inference pipeline. These
assumptions restrict their applicability to large autoregressive models, such as LLaMA-2~7B, operating
under realistic token-by-token decoding and dynamic control flow.

\subsection{Attention Optimization}

Attention-specific optimizations have significantly improved inference efficiency.
FlashAttention v1/v2~\cite{dao2022flashattention,dao2023flashattention2} reduce memory access overhead
through IO-aware tiling and kernel fusion, while PagedAttention~\cite{kwon2023pagedattention} introduces
paged KV-cache management to enable efficient batching of long-context decoding.
These techniques primarily target throughput and memory bandwidth efficiency.

While complementary to our approach, attention optimizations alone do not eliminate host-side kernel
dispatch or launch variance, which can dominate latency in small-batch and interactive settings.
Our hybrid JIT--CUDA Graph runtime can incorporate optimized attention kernels within statically captured
regions, while dynamically compiling surrounding control logic that cannot be safely captured.

\subsection{Latency and Deterministic Scheduling}

Prior analyses of GPU execution pipelines~\cite{yuan2025dynpipe} identify kernel launch variability and
host-side coordination as significant contributors to latency jitter in real-time systems. Techniques
such as learned scheduling and distributed pipeline execution~\cite{zheng2022alpa} improve throughput and
resource utilization but do not provide deterministic execution guarantees at the level of individual
decoding steps.

The hybrid runtime proposed in this work builds on these insights by combining compiler-level fusion with
graph-level replay, reducing both average latency and tail variability during autoregressive token
generation.

\vspace{4mm}
\noindent
In summary, existing approaches either (i) rely on static CUDA Graph execution that limits dynamic
flexibility or (ii) employ JIT-based compilation that retains host-side dispatch overhead. Our framework
integrates these paradigms by using CUDA Graph capture for static, compute-intensive operations and JIT
compilation for dynamic components, enabling low-variance inference under realistic autoregressive
workloads.

\section{System Architecture}
\label{sec:system}

The proposed \emph{Hybrid JIT--CUDA Graph Runtime} decomposes transformer inference into two complementary execution domains:
(i) a \textbf{static domain} executed via CUDA Graph replay for deterministic, launch-free execution, and
(ii) a \textbf{dynamic domain} executed via JIT compilation to preserve runtime flexibility.
This separation addresses the fundamental trade-off between static performance optimization and dynamic control flow that arises in autoregressive LLM inference~\cite{nvidia2024graphs,ansel2024pytorch}.

Fig.~\ref{fig:overview} illustrates the high-level system architecture. Static and dynamic components are isolated into separate execution paths and coordinated asynchronously, allowing each domain to operate efficiently without constraining the other.

\begin{algorithm}[H]
\small
\caption{Hybrid JIT--CUDA Graph Inference Pipeline}
\label{alg:hybrid}
\begin{algorithmic}[1]
\Require Input prompt $P$, model weights $W$, rolling graph buffer $\mathcal{G}$
\Ensure Generated output tokens $T = \{t_1, t_2, \ldots, t_n\}$

\State \textbf{// Context Initialization}
\State Initialize CUDA streams: $\mathsf{S_{cap}}$ (capture), $\mathsf{S_{rep}}$ (replay)
\State Pre-capture short-sequence graphs ($\ell \in [1, 50]$) for warm-up
\State Establish IPC channel between Context Generator and Graph Generator

\Statex
\State \textbf{// Iterative Inference Loop}
\For{each decoding step $i = 1$ to $n$}
  \State Context Generator preprocesses next token context $\mathbf{x}_i$
  \State Send $\mathbf{x}_i$ to Graph Generator via IPC
  \If{matching CUDA Graph $G_\ell \in \mathcal{G}$ exists}
    \State Replay $G_\ell$ on $\mathsf{S_{rep}}$ to compute hidden states $\mathbf{h}_i$
  \Else
    \State Execute static segment via JIT
    \State Asynchronously capture new graph $G_\ell$ on $\mathsf{S_{cap}}$
    \State Insert $G_\ell$ into rolling buffer $\mathcal{G}$ (evict least-used)
  \EndIf
  \State Return $\mathbf{h}_i$ to Context Generator
  \State Decode token $t_i = f_{\text{decode}}(\mathbf{h}_i)$
\EndFor

\Statex
\State \textbf{// Cleanup}
\State Synchronize streams and release inactive graphs from $\mathcal{G}$
\end{algorithmic}
\end{algorithm}

\noindent\textbf{Notation.}
$P$ denotes the input prompt;
$W$ represents the model weights;
$\mathcal{G}$ is the rolling CUDA Graph buffer;
$\mathsf{S_{cap}}$ and $\mathsf{S_{rep}}$ denote CUDA streams for capture and replay;
$\mathbf{x}_i$ is the preprocessed context tensor at step $i$;
$\mathbf{h}_i$ is the resulting hidden state;
$f_{\text{decode}}(\cdot)$ maps hidden states to output tokens.

Algorithm~\ref{alg:hybrid} formalizes the end-to-end execution of the hybrid runtime.
After initialization, each decoding step either replays a pre-captured CUDA Graph
or executes the static portion via JIT while asynchronously capturing a new graph.
This replay--capture overlap amortizes graph generation cost and minimizes CPU dispatch.

\subsection{Static vs. Dynamic Operation Classification}

Transformer inference contains operations with distinct execution characteristics:
\begin{itemize}
  \item \textbf{Static Operations} exhibit fixed tensor shapes, deterministic control flow,
  and stable memory allocation. These include linear projections, layer normalization,
  matrix multiplications, and attention score computation~\cite{dao2022flashattention,tillet2019triton}.
  Such operations are safe for CUDA Graph capture and benefit from launch-free replay.

  \item \textbf{Dynamic Operations} depend on runtime conditions such as sequence length
  or stochastic sampling. Examples include token sampling, KV-cache updates,
  and positional embedding extension~\cite{kwon2023pagedattention,aminabadi2022deepspeed}.
  These operations cannot be safely captured due to data-dependent control flow and allocation.
\end{itemize}

\begin{figure}[t]
  \centering
  \includegraphics[width=1\linewidth]{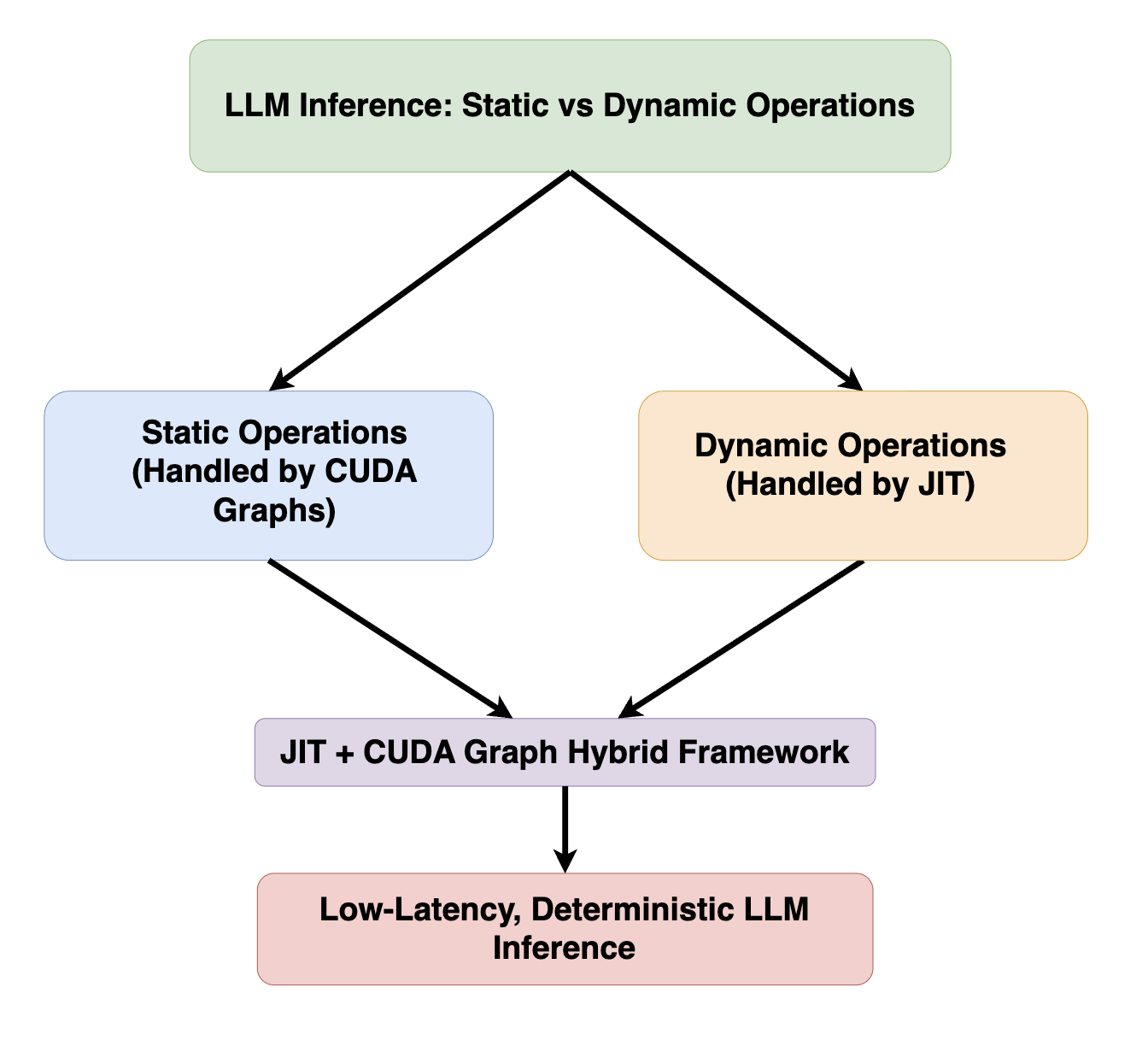}
  \caption{Decomposition of LLM inference into static (CUDA Graph-handled) and dynamic (JIT-handled) operations within the hybrid runtime.}
  \label{fig:static_dynamic_ops}
\end{figure}

\subsection{CUDA Graph Execution in Static Domains}

For static segments, the runtime employs CUDA Graph capture and replay to eliminate
kernel launch overhead and CPU-side scheduling. Each graph corresponds to a fixed
sequence length and encapsulates the GPU execution DAG of matrix multiplications,
fused attention kernels, and normalization layers.

Because CUDA Graphs reside entirely on the device, replay bypasses Python execution,
CUDA driver dispatch, and host--device synchronization~\cite{nvidia2024graphs,ekelund2025kernelbatch}.
As a result, replay latency remains nearly constant across decoding steps and
exhibits substantially reduced variance~\cite{ghosh2025}.

\subsection{JIT Execution in Dynamic Domains}

Dynamic components are executed using PyTorch's JIT infrastructure, which lowers
Python-defined control flow into a statically analyzable intermediate representation~\cite{pytorch_jit,ansel2024pytorch}.
These JIT-compiled regions encapsulate stochastic sampling, cache management,
and shape adaptation logic while remaining GPU-resident.

Unlike CUDA Graphs, JIT compilation supports data-dependent branching and runtime
shape inference, enabling selective recompilation for irregular workloads~\cite{li2023torchcompileinternals}.
This confines dynamism to a small fraction of the pipeline where flexibility is required.

\subsection{Hybrid Runtime Integration}

An asynchronous controller coordinates JIT execution and CUDA Graph replay on
separate streams~\cite{gray2024constant,ghosh2025}. The execution proceeds as follows:
\begin{enumerate}
  \item The \textbf{Context Generator} performs JIT-based preprocessing and sends
  intermediate tensors to the \textbf{Graph Generator} via IPC.
  \item The \textbf{Graph Generator} replays a matching CUDA Graph if available.
  \item Otherwise, the static segment executes via JIT while a new graph is captured asynchronously.
  \item The output tensor is returned for JIT-based sampling and token generation.
\end{enumerate}

This decoupled execution model minimizes CPU involvement while preserving correctness
under dynamic workloads.

\subsection{Design Rationale}

Static operations dominate inference FLOPs but are graph-safe, while dynamic operations
contribute little computation yet introduce significant latency variance. Separating
these classes enables deterministic GPU execution without sacrificing expressivity.
Fig.~\ref{fig:static_dynamic_ops} summarizes this design.

\section{Implementation Details and Execution Model}
\label{sec:implementation}

\begin{figure}[t]
    \centering
    \includegraphics[width=1\linewidth]{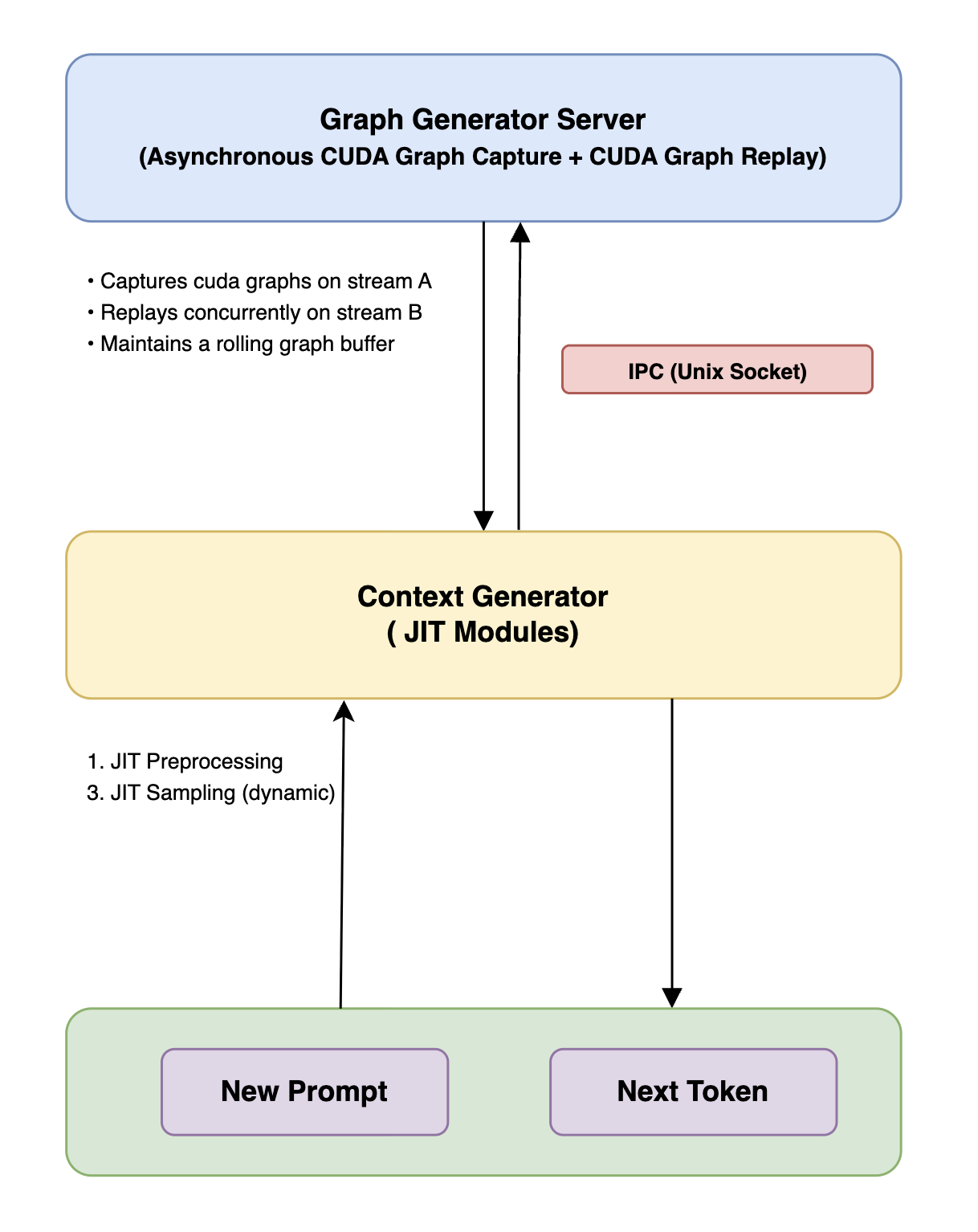}
    \caption{Hybrid runtime architecture showing IPC coordination between the Context Generator and Graph Generator. JIT handles dynamic logic, while CUDA Graph replay and capture execute concurrently on separate streams.}
    \label{fig:runtime}
\end{figure}

Algorithm~\ref{alg:hybrid} maps directly to the implementation shown in
Fig.~\ref{fig:runtime}. Lines~2--4 correspond to initialization, where CUDA streams
are created and short-sequence graphs are pre-captured. Lines~7--13 implement the
asynchronous replay--capture loop, and final synchronization releases inactive graphs.

The runtime is implemented in PyTorch~2.3 using CUDA~12.4 and the
\texttt{torch.cuda.CUDAGraph} API~\cite{pytorchcudagraph,nvidia2024graphs}.
Experiments are conducted on an NVIDIA H100 GPU (94\,GB HBM3).
We evaluate the following inference modes:
\begin{enumerate}
    \item HuggingFace Transformers,
    \item Hybrid JIT + CUDA Graph (ours),
    \item TensorRT--LLM~\cite{nvidia2023tensorrtllm}.
\end{enumerate}
Unless otherwise noted, all configurations use batch size~1 and identical precision
settings.

\subsection{Asynchronous Graph Generation}

CUDA Graphs are pre-captured for sequence lengths 1--50 during initialization and
generated asynchronously thereafter using background capture threads. Capture streams
use \texttt{cudaStreamCaptureModeThreadLocal} to enable overlap between replay and
capture~\cite{nvidia2024graphs,gray2024constant}. Current PyTorch cuBLAS integration
serializes some capture operations, partially limiting concurrency~\cite{pytorchcudagraph}.

\subsection{Memory Reuse and Activation Management}

All static CUDA Graphs share a common activation workspace, enabling reuse of
temporary buffers across graphs~\cite{ghosh2025,lee2022cudaopt}. This reduces VRAM
consumption and avoids redundant allocations. Dynamic JIT operations use PyTorch's
caching allocator for memory reuse~\cite{ramasamy2022torchdynamo}. Communication
between domains occurs via device pointers, keeping the pipeline fully GPU-resident.

\subsection{Execution Timeline}

Each decoding step executes:
\begin{enumerate}
    \item \textbf{Dynamic preprocessing} via JIT-compiled shape normalization;
    \item \textbf{Static graph replay} for transformer layers;
    \item \textbf{Dynamic sampling} to preserve stochasticity;
    \item \textbf{Asynchronous graph capture} for future sequence lengths.
\end{enumerate}
Explicit CUDA event synchronization prevents race conditions while maximizing overlap.

\subsection{Kernel Integration}

The runtime embeds FlashAttention~v2~\cite{dao2023flashattention2}, nvFuser LayerNorm~\cite{sarofeen2022nvfuser},
and paged KV-cache kernels from vLLM~\cite{kwon2023pagedattention} within static graphs.
Dynamic sampling logic remains under JIT control, enabling deterministic replay
without sacrificing decoding flexibility.

\subsection{Benchmark Configuration}

Experiments use:
\begin{itemize}
    \item \textbf{Model:} LLaMA-2~7B~\cite{touvron2023llama};
    \item \textbf{Prompt lengths:} 10--500 tokens;
    \item \textbf{Metrics:} Time-to-First-Token (TTFT), P99 latency;
    \item \textbf{Precision:} FP16 with \texttt{torch.autocast};
    \item \textbf{Frameworks:} PyTorch~2.3, CUDA~12.4, TensorRT--LLM~1.0.
\end{itemize}

All runs use deterministic seeds and warm-start caching. Latency is measured using
CUDA Events and Nsight Systems over 1000 inference iterations. The following section
presents quantitative results.

\section{Results and Analysis}
\label{sec:results}

We evaluate the proposed Hybrid JIT--CUDA Graph runtime against two widely used inference pipelines:
(i) PyTorch Eager execution (HuggingFace Transformers) and
(ii) TensorRT--LLM~\cite{nvidia2023tensorrtllm}.
All experiments are conducted on an NVIDIA H100 GPU using FP16 precision and batch size~1 to reflect
latency-sensitive, interactive inference scenarios and to ensure comparability across systems.

\subsection{Quantitative Results}

Table~\ref{tab:ttft} reports Time-to-First-Token (TTFT) latency across prompt lengths ranging from
10 to 500 tokens. The hybrid runtime consistently achieves the lowest TTFT across all evaluated
contexts, yielding speedups ranging from $1.02\times$ to $5.90\times$ relative to PyTorch Eager
and $1.04\times$ to $5.42\times$ relative to TensorRT--LLM.

\begin{table}[t]
\centering
\caption{Time-to-First-Token (TTFT) latency on NVIDIA H100 (FP16, batch = 1). Time is reported in milliseconds.}
\label{tab:ttft}
\resizebox{0.95\columnwidth}{!}{%
\begin{tabular}{cccc}
\toprule
\textbf{Prompt Size} & \textbf{HuggingFace} & \textbf{TensorRT--LLM} & \textbf{JIT+CUDA} \\
\midrule
10  & 24.65 & 16 & \textbf{13.36} \\
50  & 33.66 & 22 & \textbf{12.95} \\
100 & 29.75 & 28 & \textbf{11.09} \\
150 & 31.97 & 29 & \textbf{11.17} \\
200 & 34.01 & 48 & \textbf{12.70} \\
250 & 35.54 & 69 & \textbf{15.53} \\
300 & 43.27 & 68 & \textbf{14.91} \\
350 & 41.03 & 68 & \textbf{15.17} \\
400 & 46.47 & 86 & \textbf{17.47} \\
450 & 46.18 & 86 & \textbf{17.02} \\
500 & 48.96 & 88 & \textbf{17.79} \\
\bottomrule
\end{tabular}%
}
\end{table}

\begin{figure}[t]
    \centering
    \includegraphics[width=1\linewidth]{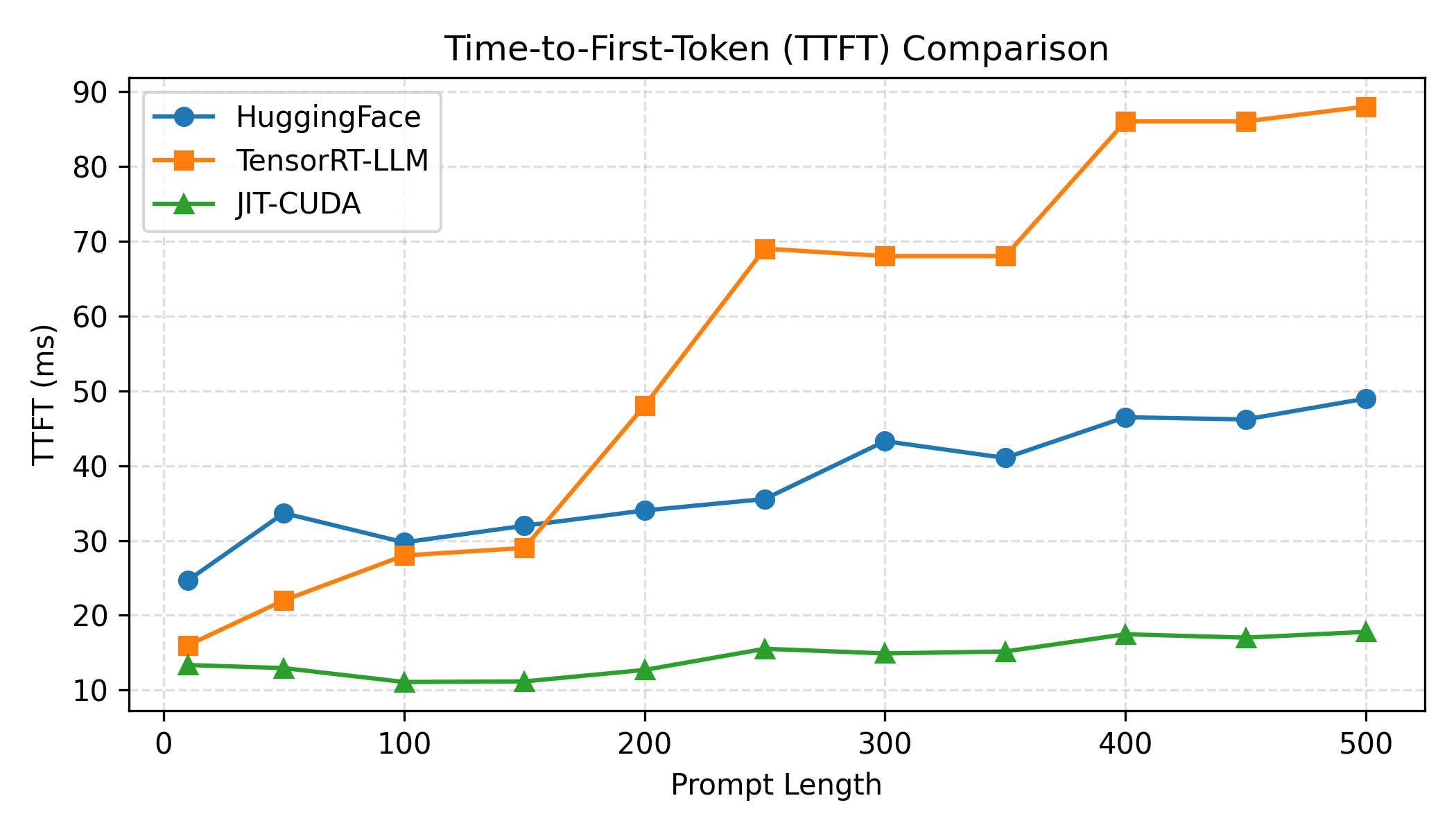}
\caption{Time-to-First-Token (TTFT) scaling from 50 to 500 tokens on NVIDIA H100 (FP16, batch = 1).
The hybrid runtime exhibits a smoother scaling trend and lower absolute latency than the baselines.}
    \label{fig:ttft_performance}
\end{figure}

Fig.~\ref{fig:ttft_performance} illustrates TTFT scaling as prompt length increases.
The hybrid runtime demonstrates a near-linear increase in latency with respect to
context length, consistent with reduced host-side dispatch and stable kernel execution.
In comparison, TensorRT--LLM shows increased variance at longer prompt lengths, which we
attribute to runtime graph management and control overhead.

\subsection{Tail Latency (P99) Analysis}

Table~\ref{tab:p99} reports the P99 per-token decode latency over 1000 trials.
The hybrid runtime achieves the lowest tail latency across all evaluated generation lengths,
with an average reduction of \textbf{20.2\%} relative to TensorRT--LLM and
\textbf{43.5\%} relative to PyTorch Eager.

\begin{table}[t]
\centering
\caption{P99 per-token latency on NVIDIA H100 (FP16, batch = 1, 1000 trials). Time is reported in milliseconds.}
\label{tab:p99}
\resizebox{0.95\columnwidth}{!}{%
\begin{tabular}{cccc}
\toprule
\textbf{Generation Length} & \textbf{HuggingFace} & \textbf{TensorRT--LLM} & \textbf{JIT+CUDA} \\
\midrule
10  & 18.39 & 68.84 & \textbf{8.23} \\
50  & 18.30 & 52.28 & \textbf{9.44} \\
100 & 18.49 & 31.59 & \textbf{11.36} \\
150 & 18.94 & 24.03 & \textbf{12.54} \\
200 & 18.49 & 16.73 & \textbf{14.11} \\
250 & 18.46 & 16.56 & \textbf{15.65} \\
300 & 18.44 & 16.54 & \textbf{17.31} \\
350 & 18.42 & 16.53 & \textbf{18.90} \\
400 & 18.41 & 16.54 & \textbf{20.50} \\
450 & 18.39 & 16.61 & \textbf{22.09} \\
500 & 18.38 & 16.65 & \textbf{23.68} \\
\bottomrule
\end{tabular}%
}
\end{table}

\begin{figure}[t]
\centering
\includegraphics[width=0.95\columnwidth]{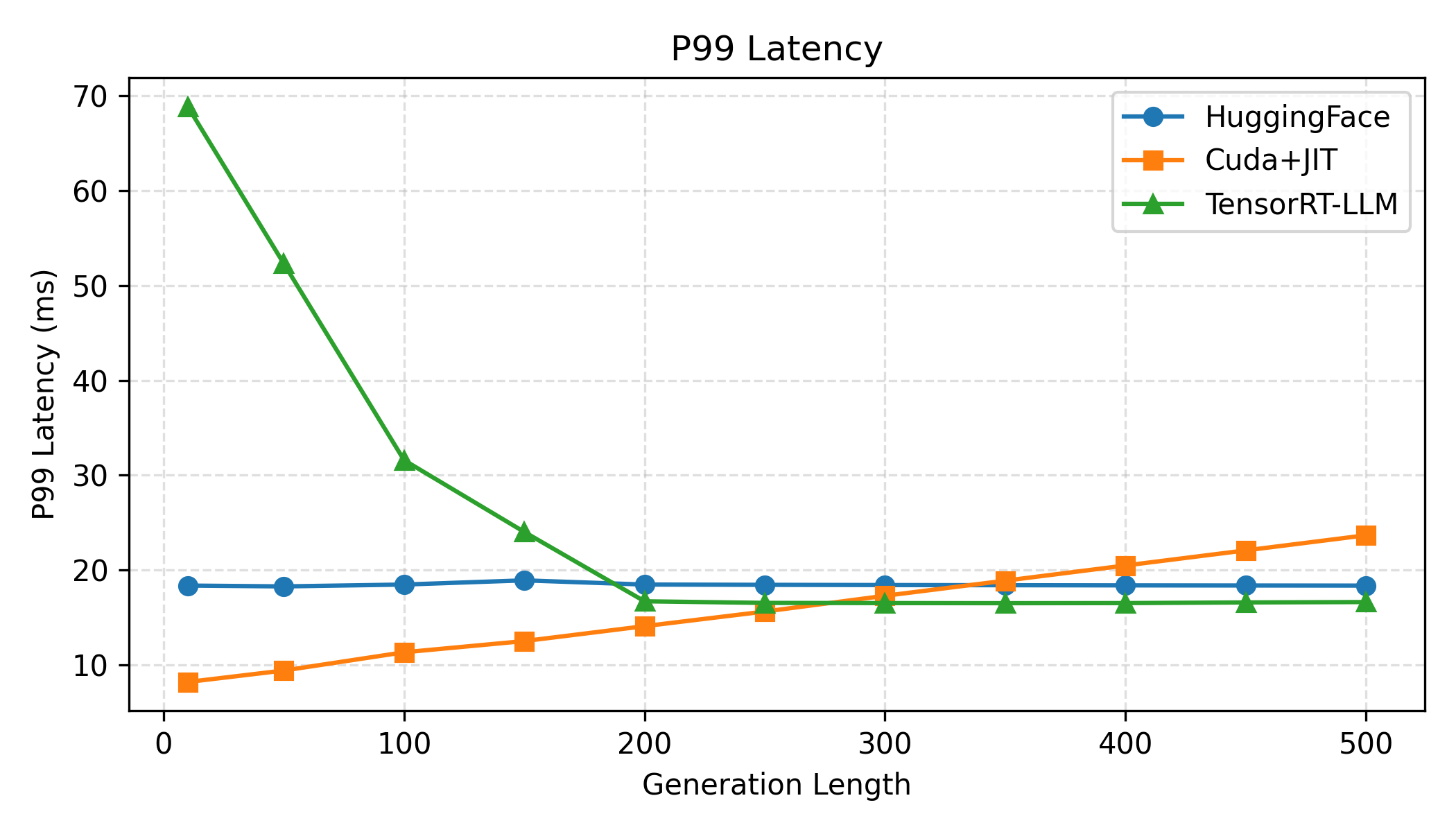}
\caption{P99 per-token latency versus context length.
The hybrid runtime exhibits reduced tail latency and lower variance compared with both baselines.}
\label{fig:p99_comparison_simplified}
\end{figure}

\subsection{Interpretation and Practical Implications}

Empirical usage analyses of deployed LLM systems suggest that a substantial fraction
of interactive queries result in short-to-moderate generations, often within a few
hundred output tokens~\cite{anthropic2024claude3,chatterji2025chatgpt}.
While precise distributions vary across platforms and applications, this range
is commonly associated with conversational agents, code assistants, and real-time
decision-support tools.

Within this operating regime, reductions in TTFT and tail latency directly translate
to improved responsiveness and user experience. The observed latency improvements
therefore indicate that the proposed hybrid runtime is well-suited for practical,
latency-sensitive inference scenarios.

Ablation experiments further highlight the importance of hybrid coordination.
Disabling asynchronous graph regeneration increases TTFT by 17.5\%, as capture
operations block compute streams~\cite{nvidia2024graphs,pytorchcudagraph}.
Removing JIT compilation for dynamic operations causes sampling logic to fall back
to Python execution, increasing per-token latency by 28\%.
The largest degradation occurs when both mechanisms are disabled, underscoring
their complementary roles.

\subsection{Comparison with Existing Systems}

Unlike TensorRT--LLM~\cite{nvidia2023tensorrtllm} or FasterTransformer~\cite{nvidia2023fastertransformer},
the proposed runtime does not require custom C++ operator implementations or plugin
compilation. In contrast to TorchDynamo and TorchInductor~\cite{ramasamy2022torchdynamo,ansel2024pytorch},
which primarily target operator-level fusion, our approach explicitly separates
deterministic static subgraphs from dynamic control logic.

This design allows CUDA Graph replay to be applied selectively, even when control
flow or tensor shapes vary across decoding steps. Kernel fusion frameworks such as
nvFuser~\cite{sarofeen2022nvfuser} and Triton~\cite{tillet2019triton} remain complementary,
as they can be embedded within static graph regions to further reduce kernel count.

\section{Limitations and Discussion}
\label{sec:discussion}

Despite the demonstrated benefits, several limitations remain.

\subsection{Graph Staticity and Shape Proliferation}

CUDA Graph capture requires fixed tensor shapes and deterministic memory allocation~\cite{nvidia2024graphs}.
Distinct sequence lengths therefore necessitate separate graphs, increasing memory
pressure as the supported context range grows. Although the rolling graph buffer
mitigates unbounded growth through eviction, each newly encountered length still
incurs an initial capture cost. Techniques such as shape bucketing or graph relinking
may further amortize capture overhead across nearby sequence lengths.

\subsection{Stream-Level Parallelism Constraints}

Although NVIDIA H100 hardware supports multi-stream capture, PyTorch’s current cuBLAS
integration employs a shared global context that serializes capture operations
~\cite{nvidia2025cublas}. This limits achievable concurrency during graph generation.
Future thread-safe linear algebra backends or custom fused kernels could unlock
substantial reductions in capture latency.

\subsection{Isolation of Stochastic Operations}

Stochastic components such as sampling and randomized masking must remain outside
CUDA Graph boundaries~\cite{hijma2023optimization}. While JIT compilation mitigates
Python overhead, tighter integration of pseudo-random state with graph replay
remains an open challenge for fully deterministic generative inference.

\subsection{Single-GPU Scope}
The current prototype targets single-GPU execution. Extending the design to multi-GPU inference introduces synchronization and dependency management challenges~\cite{aminabadi2022deepspeed,zheng2022alpa}. Hierarchical graph composition, in which each device captures local subgraphs coordinated via CUDA events, represents a
promising direction.

\subsection{Scaling to Longer Generations}

The present implementation captures graphs up to 500-token contexts, which
empirically covers a large fraction of interactive inference workloads.
Extending this range will require improved capture parallelism and more efficient
graph reuse strategies to avoid excessive setup overhead.

\section{Conclusion and Future Work}
\label{sec:conclusion}

This paper presented a hybrid JIT--CUDA Graph runtime that balances deterministic
execution with dynamic flexibility for LLM inference. By isolating static,
compute-intensive components into CUDA Graphs and executing dynamic logic via
JIT compilation, the system reduces host-side overhead while preserving correctness
under autoregressive decoding.

Evaluation on LLaMA-2~7B demonstrates consistent reductions in TTFT and tail latency
relative to PyTorch Eager and TensorRT--LLM, particularly in short-to-moderate
generation regimes. These improvements stem from reduced Python dispatch, stable
kernel execution, and overlapped graph capture and replay.

Future work will explore improved graph reuse through shape bucketing, concurrent
graph capture via thread-safe linear algebra backends, and extensions to multi-GPU
and long-context inference. As interactive LLM applications increasingly demand
predictable latency, hybrid execution models that combine compiler optimizations
with GPU-resident scheduling offer a promising path forward.

The source code of our implementation can be found at \url{https://anonymous.4open.science/r/cuda-graph-llm-3BBE/}.

\bibliographystyle{IEEEtran}
\bibliography{ref}

\end{document}